\documentclass[copyright,creativecommons]{eptcs}
\providecommand{\event}{Technical Communications of ICLP 2025} 

\usepackage{iftex}

\ifpdf
  \usepackage{underscore}         \usepackage[T1]{fontenc}        \else
  \usepackage{breakurl}           \fi

\usepackage[utf8]{inputenc}
\usepackage{amsmath}
\usepackage{amssymb}
\usepackage{microtype}
\usepackage{url}
\usepackage{graphicx}
\usepackage{xcolor}
\usepackage{subcaption}
\usepackage{wrapfig}

\usepackage{listings}
\providecommand{\Underscore}{\textunderscore}

\lstdefinelanguage{clingo}{basicstyle=\ttfamily,keywordstyle=[1]\bfseries,keywordstyle=[2]\bfseries,keywordstyle=[3]\bfseries,showstringspaces=false,literate={_}{\Underscore}1 {\%\%}{}0,escapeinside={\#(}{\#)},alsoletter={\#,\&},keywords=[1]{not,from,import,def,if,else,elif,return,while,break,and,or,for,in,del,and,class,with,as,is,yield,async},keywords=[2]{\#const,\#show,\#minimize,\#base,\#theory,\#count,\#external,\#program,\#script,\#end,\#heuristic,\#edge,\#project,\#show,\#sum},morecomment=[l]{\#\ },morecomment=[l]{\%\ },morestring=[b]",stringstyle={\itshape},commentstyle={\color{darkgray}}}

\lstdefinelanguage{clingcon}[]{clingo}{morekeywords={&dom,&sum,&nsum,&diff,&disjoint,&distinct,&minimize,&maximize,&show}}
\lstdefinelanguage{fclingo}[]{clingo}{morekeywords={&sum,&sus,&in,&df,&min,&max,&show}}
\lstdefinelanguage{clingodl}[]{clingo}{morekeywords={&diff}}

\lstdefinelanguage{python}{basicstyle=\ttfamily,keywordstyle=[1]\bfseries,showstringspaces=false,literate={_}{\Underscore}{1},escapeinside={\#(}{\#)},alsoletter={\#,\&},keywords=[1]{not,from,import,def,if,else,elif,return,while,break,and,or,for,in,del,and,class,with,as,is,yield,async},morecomment=[l]{\#\ },morestring=[b]",stringstyle={\itshape},commentstyle={\color{darkgray}}}

 \lstdefinelanguage{clingos}{language=clingo,basicstyle=\small\ttfamily }
\lstdefinelanguage{clingof}{language=clingo,basicstyle=\footnotesize\ttfamily }

\definecolor{aol-bg}{HTML}{FAFAFA}
\definecolor{aol-blue}{HTML}{4078F2}   \definecolor{aol-navy}{HTML}{36597A}   \definecolor{aol-green}{HTML}{50A14F}  \definecolor{aol-gray}{HTML}{A0A1A7}   \definecolor{aol-purple}{HTML}{A626A4} \definecolor{aol-orange}{HTML}{E45649} 

\lstdefinelanguage{clingof-color}{
  language=clingo,
  basicstyle=\footnotesize\ttfamily\color{aol-blue},
  numberstyle=\footnotesize\color{black},
  rulecolor=\color{black},
  stringstyle=\color{aol-green},
  commentstyle=\color{gray},
  morekeywords=[1]{\#minimize, \#maximize, \#sum, \#const, \#show, \#program},
  keywordstyle=[1]\color{aol-purple},
  morekeywords=[2]{not},
  keywordstyle=[2]\color{aol-purple},
  literate={:-}{{\textcolor{black}{\texttt{:-}}}}2
           {\{}{{\textcolor{black}{\texttt{\{}}}}1
           {\}}{{\textcolor{black}{\texttt{\}}}}}1
           {[}{{\textcolor{black}{\texttt{[}}}}1
           {]}{{\textcolor{black}{\texttt{]}}}}1
           {(}{{\textcolor{aol-navy}{\texttt{(}}}}1
           {)}{{\textcolor{aol-navy}{\texttt{)}}}}1
           {,}{{\textcolor{black}{\texttt{,}}}}1
           {.}{{\textcolor{black}{\texttt{.}}}}1
           {:}{{\textcolor{black}{\texttt{:}}}}1
           {;}{{\textcolor{black}{\texttt{;}}}}1
           {=}{{\textcolor{black}{\texttt{=}}}}1
           {>}{{\textcolor{black}{\texttt{>}}}}1
           {<}{{\textcolor{black}{\texttt{<}}}}1
           {-}{{\textcolor{black}{\texttt{-}}}}1
           {*}{{\textcolor{black}{\texttt{*}}}}1
}

\makeatletter\lst@AddToHook{OnEmptyLine}{\vspace{\dimexpr-\baselineskip+\smallskipamount\relax}}\makeatother
\providecommand{\sysfont}{\textit}

\newcommand{\clingo}{\sysfont{clingo}}

\newcommand{\sysname}{{\sysfont{aspital}}}

\newenvironment{ourdescription}[0]{\par}{\par}
\newcommand{\ouritem}[1]{\par\noindent{\textbf{#1}}}
\newcommand{\citep}[1]{\cite{#1}}
\newcommand{\citealp}[1]{\cite{#1}}
\title{The ASP-based Nurse Scheduling System at the University of Yamanashi Hospital}
\author{
  {Hidetomo} {Nabeshima}
  \institute{University of Yamanashi, Japan}
  \and
  {Mutsunori} {Banbara}
  \institute{Nagoya University, Japan}
  \and
  {Torsten} {Schaub}
  \institute{University of Potsdam, Germany}
  \and
  {Takehide} {Soh}
  \institute{Nagoya University, Japan}
}

\newcommand{\titlerunning}{Nurse Scheduling System}
\newcommand{\authorrunning}{Nabeshima et al.}

\hypersetup{
  bookmarksnumbered,
  pdftitle    = {\titlerunning},
  pdfauthor   = {\authorrunning},
  pdfsubject  = {EPTCS},
  pdfkeywords = {Answer Set Programming, Nurse Scheduling}
}

\makeatletter
\begin{document}
\maketitle
\begin{abstract}
We present the design principles of a nurse scheduling system
built using Answer Set Programming (ASP) and
successfully deployed at the University of Yamanashi Hospital.
Nurse scheduling is a complex optimization problem requiring the reconciliation of individual nurse preferences with
hospital staffing needs across various wards.
This involves balancing hard and soft constraints and the flexibility of interactive adjustments.
While extensively studied in academia,
real-world nurse scheduling presents unique challenges that go beyond typical benchmark problems and competitions.
This paper details the practical application of ASP to address these challenges at the University of Yamanashi Hospital,
focusing on the insights gained and the advancements in ASP technology necessary to effectively manage the complexities of
real-world deployment.
\end{abstract}
\section{Introduction}\label{sec:introduction}
The University of Yamanashi
Hospital\footnote{\url{https://www.hosp.yamanashi.ac.jp}} serves as a core
medical institution providing advanced medical care in the region and is one of
the largest hospitals in the Yamanashi prefecture. It has approximately 620 beds,
400 physicians, and 760 nurses.
The hospital consists of 19 wards, each with its own work schedule for the nurses
assigned to it.

In our hospital, nurse schedules are manually created by head nurses, requiring
significant effort and time. To automate the scheduling process, we formulated
the Nurse Scheduling Problem (NSP) specific to our hospital.
We first presented a typical NSP formulation for Japanese hospitals
\citep{ikegami2005} to the head nurses. Based on their feedback, we refined the
model by adding missing entities or constraints and removing redundant ones,
thereby developing a base model.
Subsequently, we focused on the obstetrics and gynecology wards, which have
particularly strict requirements. With our NSP model, we automatically generated
schedules for this ward. These schedules were then evaluated by the head nurse
of the ward, who provided feedback by identifying unacceptable or unnatural
elements.  Based on this feedback, we refined the formulation of the NSP.
We call this process the \emph{iterative modeling refinement cycle}, which
involves generating schedules, evaluating the results, and refining the
formulation iteratively.
This cycle was repeated over approximately one year, during which schedules were
generated and refined every four weeks.
The NSP formulation presented in this section represents the current state of
the model, which has been tailored to meet the specific needs of our hospital.

Head nurses possess both explicit and implicit knowledge regarding nurse
scheduling. While much of the explicit knowledge can be identified during the
initial formulation of the NSP, implicit knowledge often emerges during the
evaluation of automatically generated schedules. Therefore, the iterative
modeling refinement cycle is essential for extracting implicit knowledge.
However, fully capturing implicit knowledge remains a significant challenge. To
address this, we acknowledge that our NSP formulation is inherently incomplete and
propose a system to support modifications to the generated schedules,  ensuring
they better align with the head nurse's requirements. Details of this support
system are provided in the next section.

The basic entities in our NSP model include \emph{nurses}, \emph{nurse groups},
\emph{shifts}, \emph{shift groups}, \emph{dates}, \emph{past shifts}, and
\emph{requested shifts} for the current and following months.\footnote{Since the
scheduling period is divided into four-week intervals, the terms ``current'' and
``following months'' refer to these four-week units.}
Nurses are characterized by their ID and skill level.
Each nurse is assigned to exactly one ward.
The number of nurses varies by ward, ranging from 15 to 40.\footnote{We
focus on full-time nurses because part-time nurses in our hospital follow
fixed work patterns.} However, this number may fluctuate  during the scheduling
period due to factors such as new hires, retirements, or staff transfers.
Nurse groups are sets of nurses classified based on their skills and
experience.\footnote{Nurse groups are one of the commonly used entities in NSP
and correspond to ``skills'' in the INRC (The International Nurse Rostering
Competition, \url{https://mobiz.vives.be/inrc2})~\citep{ceschia19}.} Examples of
such groups include senior, intermediate, and novice groups. Nurses may belong
to multiple groups. For instance, head nurses and deputy head nurses may be part
of both the senior and leadership groups.

\begin{table}
  \centering
  \caption{Nurse Shifts Targeted for Automatic Scheduling in the Hospital}\label{tbl:shifts}
  {\begin{tabular}{@{\extracolsep{\fill}}llc}
    \hline
    Sign  & Description         & Work hours    \\ \hline
    EM    & Early morning shift & 06:00--14:45  \\
    D     & Day shift           & 08:00--16:45  \\
    LD    & Long day shift      & 08:00--20:15  \\
    LM    & Late morning shift  & 11:30--20:15  \\
    E     & Evening shift       & 16:00--00:45  \\
    SE    & Short evening shift & 19:45--00:00  \\
    N     & Night shift         & 00:00--08:45  \\
    SN    & Short night shift   & 00:00--08:45  \\
    WO    & Weekly day off      & ---           \\
    PH    & Public holiday leave& ---           \\ \hline
  \end{tabular}}
\end{table}
Table~\ref{tbl:shifts} lists the shifts subject to automatic assignment,
categorized into \emph{work shifts} and \emph{rest shifts}. Work shifts include
scheduled hours such as day, evening, and night shifts, while rest shifts
represent non-working days, including weekly rest and designated off days. In
this study, rest days are also treated as shifts.
In practice, the working hours of shifts vary across wards in the hospital,
resulting in a larger number of distinct shifts. However, in this study, these
variations are abstracted and consolidated into eight standardized work shifts
for simplicity. The working hours shown in Table~\ref{tbl:shifts} represent
typical time ranges.
Standard NSP models typically include only day, evening, and night shifts. The
inclusion of a larger variety of work shifts is one of the unique features of
our NSP model.
Additionally, there are two specific categories of shifts: \emph{duty shifts}
and \emph{leave shifts}. Duty shifts represent special assignments such as
business trips and training, while leave shifts cover various types of leave,
including annual leave, maternity leave, and childcare leave. These shifts are
assigned based on nurse requests and are not subject to automatic assignment.
Therefore, they are not included in the table.
Although N and SN have the same working hours, they are treated as separate
shifts because they are typically assigned to different groups of nurses. This
distinction is also reflected in shift patterns; for example, the SE shift is
followed by SN, while the E shift is followed by N.

A shift group is a set of shifts. For example, the shift group \{E, SE\}  is
used to specify the required number of nurses for evening shifts. Constraints
related to the number of nurses or the number of shift assignments are often
defined on shift groups. This concept of shift groups is also a unique feature
of our NSP model.

Dates include past shift dates, the current month, and the first week of the
next month to ensure consistency across scheduling boundaries.
Past shifts refer to schedules from periods prior to the current month. These
schedules are necessary to ensure consistency between assignments at the end of
the previous month and the start of the current month, and they are essential
for maintaining equitable shift allocations.
Requested shifts are classified as desired or undesired, and represented as a
triplet of nurse, date, and shift. Shifts cover work, weekly rest, leave, and
duty shifts. Including early next-month shifts helps preserve scheduling
continuity.
Notably, in our hospital, requested shifts are treated primarily as hard constraints.

The following is the list of constraints in our NSP model. Details of some
constraints are presented later through their ASP encodings. The full ASP
encoding for all constraints is available in the repository at
\url{https://github.com/nabesima/yamanashi-nsp}. This repository also includes
anonymized real instances, artificially generated instances of various sizes,
and scripts for solving the problem.
\begin{ourdescription}
  \ouritem{Workdays ($H_1$):} Nurses must be scheduled to work a specific
    number of days per scheduling period, typically 20 days. This number may
    decrease due to external factors such as business trips, public holidays, or
    paid leave.
  \ouritem{Weekly Rest Days ($H_2$):} Nurses must be assigned a specified
    number of weekly rest days, usually 8 days per period.
  \ouritem{Requested shifts ($H_3$):} Nurses must be assigned to their
  desired shifts while avoiding undesired shifts.
  \ouritem{Average Working Hours ($H_4$):} The assignments for LD and SE
    shifts must be equalized to ensure an average daily working time of 7 hours
    and 45 minutes.
  \ouritem{Consecutive Workdays ($H_5$, $S_1$):} Consecutive workdays
  beyond the specified limit, typically 5 or 6 days, are prohibited.
  \ouritem{Daily Staffing ($H_6$, $S_2$):} The required number of nurses
    must be ensured for each nurse group, each shift group, and each day.
    Alternatively, nurses must be assigned such that the total skill levels
    exceed the specified threshold.
  \ouritem{Shift Frequency ($H_7$, $S_3$):} Nurses are assigned shifts
    within the predefined range for each shift group to ensure balanced workloads.
    For example, nurses in the night shift group may be assigned 4--8 night shifts per period.
  \ouritem{Shift Patterns ($H_8$, $S_4$):} Work shift patterns must fall
    within the specified range, such as avoiding overly irregular schedules.
    For example, the shift pattern LD-D (a long day shift followed by a day shift)
    may be allowed at most once per period.
  \ouritem{Inter-Shift Rules ($H_9$, $S_5$):} Consecutive shifts must
  comply with the specified constraints. For instance, an evening shift must
  be followed by a night shift. It is recommended that the next
  shift begins at least 24 hours after the start of the previous shift.
  \ouritem{Nurse Pairing ($H_{10}$, $S_6$):} Recommended nurse pairs (e.g.,
    mentor-novice pairs) should ideally be assigned the same shifts whenever
    possible, while prohibited pairs must not work the same shifts.
  \ouritem{Isolated Workdays ($S_7$):} Isolated single workdays surrounded
    by rest days should be minimized to reduce unnecessary disruptions.
  \ouritem{Leave-Adjacent Rest Days ($S_8$):} Weekly rest days should be
    scheduled before or after requested leave days to extend rest periods.
  \ouritem{Equal Workload Distribution ($S_9$):} Shift workloads should be
    distributed evenly among nurses to ensure fairness. For example, this includes
    the number of rest days on weekends or public holidays.
\end{ourdescription}
These constraints are classified into \emph{hard constraints} ($H_i$), which
must always be satisfied, and \emph{soft constraints} ($S_i$), which should be
satisfied as much as possible. Hard constraints such as $H_1$, $H_2$, and
$H_5$–$H_8$ enforce strict lower and/or upper bounds, while soft constraints
such as $S_1$–$S_4$ and $S_6$ define desirable limits.
Violating a soft constraint incurs a penalty proportional to the square of the
deviation from the threshold to discourage large violations. Other soft
constraint violations incur a fixed penalty per occurrence.
A \emph{feasible solution} is a shift assignment to each nurse and each day that
satisfies all hard constraints. Soft constraints are prioritized, with their
importance varying by ward. The objective is to find a feasible solution that
minimizes penalty costs in lexicographical order based on these priorities.
Additionally, some wards have specific constraints. For example, in the ICU, six
consecutive workdays are allowed, but new hires are limited to five, and their
night shift assignments must be equal.

\section{ASP Encoding of Constraints}

The ASP encoding of our NSP model consists of approximately
100 rules. Due to space limitations, only a subset of the fundamental
constraints is presented here. The complete encoding is available in our
repository.
Listing~\ref{lst:basic-encoding} shows the ASP encoding for constraints
$H_1$--$H_7$ and $S_2$--$S_3$, with certain constraints omitted for brevity.
Some predicates appearing in the figure are either directly defined in the
instance or generated during preprocessing.
\begin{ourdescription}
\ouritem{Workdays} ($H_1$, lines 2--7):
Fact \verb|staff(N)| specifies that \verb|N| is a nurse, and
  \verb|work_shift(S)| indicates that \verb|S| is a work shift. Similarly,
  \verb|workable_date(N,D)| indicates that \verb|N| is available to work on day
  \verb|D|, meaning \verb|N| is not assigned to a duty shift on that day.
The predicates \verb|assigned(N,D)| and \verb|assigned(N,D,S)| define the
  schedule, with the former indicating that \verb|N| works on \verb|D| and the
  latter specifying the assigned shift \verb|S|. Line 2 ensures assignments only
  on \verb|workable_date(N,D)|, while line 3 enforces exactly one shift per
  workday. Lines 4--7 constrain the number of workdays within the bounds
  \verb|LB| and \verb|UB| specified by \verb|work_days_bounds(N,LB,UB)|.
Constraint violations are tracked using \verb|violation(T,C,LIM,VAL)|, where
  \verb|T| is the violation type (\verb|hard| or \verb|soft|), \verb|C| is
  the reason, and \verb|LIM| and \verb|VAL| denote the target and actual values.
  This predicate is discussed below.

\ouritem{Weekly Rest Days} ($H_2$, lines 10--13): These rules ensure that the
  number of weekly rest days assigned to each nurse \verb|N| falls within the
  range specified by \verb|weekly_rest_bounds(N,LB,UB)|. The fact
  \verb|weekly_rest_available_date(N,D)| indicates that \verb|N| can be assigned
  a weekly rest day on \verb|D|, provided \verb|D| is neither a public holiday
  nor a day with a paid leave request.

\ouritem{Requested Shifts} ($H_3$, lines 16--19):
The fact \verb|pos_request(N,D,S)| represents that nurse \verb|N| has a
  desired shift \verb|S| on day \verb|D|, and multiple desired shifts can be
  specified for the same day. \verb|pos_request(N,D)| denotes that \verb|N| has
  at least one desired shift on \verb|D|.
The predicate \verb|ext_assigned(N,D,S)| extends \verb|assigned(N,D,S)| to
  cover all shift types, not just work shifts. Details are explained shortly.
Lines 16--17 ensure at least one desired shift is assigned. Similarly,
  \verb|neg_request(N,D,S)| represents shifts that nurse \verb|N| prefers to avoid, and
  lines 18--19 ensure they are not assigned.
The predicate \verb|violation(T,C)| represents a constraint violation with a
  fixed penalty, while its four-argument version includes additional parameters
  for violation severity.

  The predicate \verb|ext_assigned(N,D,S)| is defined on lines 22--26. If
  \verb|assigned(N,D,S)| holds, then \verb|ext_assigned(N,D,S)| also holds (line
  22). If nurse \verb|N| has no assigned work shift on day \verb|D|, one of the
  following shifts is assigned: (1) the weekly rest shift, (2) the public
  holiday shift, or (3) a duty or leave shift requested by the nurse. The facts
  \verb|weekly_rest_avail_date(N,D)| and \verb|pub_holiday_avail_date(N,D)|
  indicate that nurse \verb|N| has no duty or leave requests on day \verb|D| and
  that the day is either a non-holiday or a holiday, respectively. The former
  corresponds to (1), where the weekly rest shift (WR) is assigned, while the
  latter corresponds to (2), assigning the public holiday shift (PH) (lines
  22--23).
The fact \verb|manual_request(N,D,S)| represents a nurse's request for a duty
  or leave shift on day \verb|D|, corresponding to (3). In this case, exactly
  one of the requested duty or leave shifts is assigned (lines 25--26).

\ouritem{Average Working Hours} ($H_4$, lines 29--30): This rule enforces that
  subtracting the number of SE shifts from the number of LD shifts assigned to
  each nurse \verb|N| results in zero, ensuring that the two shift types are
  assigned equally.

\ouritem{Consecutive Workdays} ($H_5, S_1$, lines 33--38): The fact
  \verb|consecutive_work_ub(T,| \verb|NG,UB)| specifies the upper bound \verb|UB| on
  consecutive workdays for nurse group \verb|NG|, where \verb|T| denotes the
  constraint type (\verb|hard| or \verb|soft|). In our NSP model, no lower bound is
  imposed. The fact \verb|staff_group(NG,N)| indicates that nurse \verb|N|
  belongs to nurse group \verb|NG|, and \verb|base_date(D)| includes days in the
  current month and one week before and after, accounting for consecutive work
  periods spanning adjacent months. The predicate \verb|work_day(N,D)| indicates
  that \verb|N| has either a work shift or a duty shift on day \verb|D|.
Lines 33--35 define the predicate \verb|full_work_period(N,BD,ED)| when a
  nurse works exactly up to the upper bound of consecutive workdays, with
  \verb|BD| and \verb|ED| as the start and end dates. Lines 36--38 identify a
  violation if the upper bound is exceeded by one day. If exceeded by $n$ days,
  $n$ violations are recorded.

\ouritem{Daily Staffing} ($H_6, S_2$, lines 41--50):
The fact \verb|shift_group(SG,S)| specifies that shift \verb|S| belongs to
  shift group \verb|SG|, and \verb|staff_lb(T,NG,SG,D,LB)| gives the lower
  bound \verb|LB| on the number of nurses from group \verb|NG| required for
  shift group \verb|SG| on day \verb|D|.  Lines 41--42 define violations when
  this bound is unmet. The same applies to upper bounds.
The fact \verb|point_lb(T,NG,SG,D,LB)| gives the lower bound on the total
  skill levels of nurses, and \verb|point(N,P)| specifies that nurse \verb|N| has skill
  level \verb|P|. Lines 45--47 define violations when total skill levels fall
  below \verb|LB|. Upper bounds are handled similarly.
Some wards impose constraints solely on nurse headcount, while others enforce
  both headcount and skill-level constraints. In the latter case, satisfying
  either constraint is sufficient. This is implemented by defining a rule that
  treats a violation as occurring only when both constraints are violated
  simultaneously. The detailed encoding is omitted due to space limitations.

\ouritem{Shift Frequency} ($H_7, S_3$, lines 53--56): The fact
  \verb|shift_lb(T,N,SG,LB)| specifies that the total number of shifts assigned
  to nurse \verb|N| from shift group \verb|SG| must be at least \verb|LB|, where
  \verb|T| denotes the constraint type. Lines 53--54 ensure that this total
  meets or exceeds \verb|LB|. Upper bounds are enforced similarly.
\end{ourdescription}

\begin{lstlisting}[float,language=clingof-color,frame=lines,caption={ASP Encoding of Basic Constraints},label={lst:basic-encoding}]
% H1. Workdays
{ assigned(N,D) : workable_date(N,D) } :- staff(N).
1 { assigned(N,D,S) : work_shift(S) } 1 :- assigned(N,D).
violation(hard,work_days_lb(N),LB,X) :-
    staff(N), work_days_bounds(N,LB,UB), X = { assigned(N,D) }, X < LB.
violation(hard,work_days_ub(N),UB,X) :-
    staff(N), work_days_bounds(N,LB,UB), X = { assigned(N,D) }, UB < X.

% H2. Weekly Rest Days
violation(hard,weekly_rest_lb(N),LB,X) :- weekly_rest_bounds(N,LB,UB),
    X = { not assigned(N,D) : weekly_rest_available_date(N,D) }, X < LB.
violation(hard,weekly_rest_ub(N),UB,X) :- weekly_rest_bounds(N,LB,UB),
    X = { not assigned(N,D) : weekly_rest_available_date(N,D) }, UB < X.

% H3. Requested Shifts
violation(hard,pos_request(N,D)) :- pos_request(N,D), date(D),
    not 1 { ext_assigned(N,D,S) : pos_request(N,D,S) } 1.
violation(hard,neg_request(N,D)) :- neg_request(N,D), date(D),
    1 { ext_assigned(N,D,S) : neg_request(N,D,S) }.

% Extends assigned/3 to ext_assigned/3
ext_assigned(N,D,S) :- assigned(N,D,S).
ext_assigned(N,D,"WR") :- not assigned(N,D), weekly_rest_avail_date(N,D).
ext_assigned(N,D,"PH") :- not assigned(N,D), pub_holiday_avail_date(N,D).
1 { ext_assigned(N,D,S) : manual_request(N,D,S) } 1 :-
    not assigned(N,D), manual_request(N,D).

% H4. Average Working Hours
violation(hard,eq_shifts(N)) :- staff(N),
    not 0 #sum{ 1,D : assigned(N,D,"LD") ; -1,D : assigned(N,D,"SE") } 0.

% H5. Consecutive Workdays
full_work_period(N,BD,ED) :- consecutive_work_ub(_,NG,UB), staff(N),
    staff_group(NG,N), base_date(BD), ED=BD+UB-1, base_date(ED), -1 <= ED,
    work_day(N,D) : D = BD..ED.
violation(T,consecutive_work_days(N, BD),UB,UB+1) :-
    consecutive_work_ub(T,NG,UB), staff_group(NG,N),
    full_work_period(N,BD,ED), work_day(N,ED+1).

% H6, S2. Daily Staffing
violation(T,staff_lb(NG,SG,D),LB,X) :- staff_lb(T,NG,SG,D,LB),
    X = { assigned(N,D,S) : staff_group(NG,N), shift_group(SG,S) }, X < LB.
violation(T,staff_ub(NG,SG,D),UB,X) :- staff_ub(T,NG,SG,D,UB),
    X = { assigned(N,D,S) : staff_group(NG,N), shift_group(SG,S) }, UB < X.
violation(T,point_lb(NG,SG,D),LB,X) :- point_lb(T,NG,SG,D,LB),
    X = #sum{ P,N : point(N,P), assigned(N,D,S),
                    staff_group(NG,N), shift_group(SG,S) }, X < LB.
violation(T,point_ub(NG,SG,D),UB,X) :- point_ub(T,NG,SG,D,UB),
    X = #sum{ P,N : point(N,P), assigned(N,D,S),
                    staff_group(NG,N), shift_group(SG,S) }, UB < X.

% H7, S3. Shift Frequency
violation(T,shift_lb(N,SG),LB,X) :- shift_lb(T,N,SG,LB),
    X = { assigned(N,D,S) : shift_group(SG,S) }, X < LB.
violation(T,shift_ub(N,SG),UB,X) :- shift_ub(T,N,SG,UB),
    X = { assigned(N,D,S) : shift_group(SG,S) }, UB < X.
\end{lstlisting}
Listing~\ref{lst:violations} defines the treatment of constraint violations and the
objective function.
The predicate \verb|soften_hard| is used to relax hard constraints.
Hospital wards often face nursing staff shortages due to scheduled absences
(e.g., training, business trips) and unexpected long-term absences (e.g.,
illness, injury). Additionally, idealistic staffing requirements set by head
nurses can further restrict shift flexibility.
These factors frequently lead to constraint violations, making it difficult to
fully adhere to hard constraints.
To address this, a mechanism for relaxing hard constraints is essential.
By default, \verb|soften_hard| is set to false. If the NSP instance is
unsatisfiable, enabling it allows schedule generation despite hard constraint
violations.
The first line defines its choice rule, with its truth value
provided as an assumption in incremental ASP solving \citep{karoscwa21a}.
This mechanism enhances response time by allowing the solver to resume search
immediately instead of redoing both grounding and search when toggling
\verb|soften_hard|.
Lines 2 and 3 enforce that hard constraint violations are prohibited unless
\verb|soften_hard| is true.

Lines 5--7 encode penalty calculations for constraint violations. The predicate
\verb|penalty(T,C,W,P)| represents a penalty, where \verb|T| denotes the
constraint type, \verb|C| the penalty reason, \verb|W| the weight, and \verb|P|
the priority. The fact \verb|priority(T,C,P)| assigns priority \verb|P| to
reason \verb|C| with the constraint type \verb|T|. If a violation has a severity
level, its squared value is used as the penalty weight (lines 5--6); otherwise,
a constant weight is applied (line 7).
The objective function, defined on line 9, minimizes the sum of penalty weights
in lexicographical order based on priority.

\begin{lstlisting}[float,language=clingof-color,frame=lines,aboveskip=10pt,label={lst:violations},caption={ASP Encoding for Constraint Violations and Objective Function.}]
{ soften_hard }.
:- violation(hard,_,_,_), not soften_hard.
:- violation(hard,_), not soften_hard.

penalty(T,C,W,P) :- violation(T,C,LIM,VAL),
  W = (LIM-VAL)*(LIM-VAL), priority(T,C,P).
penalty(T,C,1,P) :- violation(T,C), priority(T,C,P).

#minimize { W@P,T,C : penalty(T,C,W,P) }.
\end{lstlisting}

\section{Nurse Scheduling System}
\label{sec:system}
\begin{figure}[tb]
  \centering
  \includegraphics[width=0.8\textwidth]{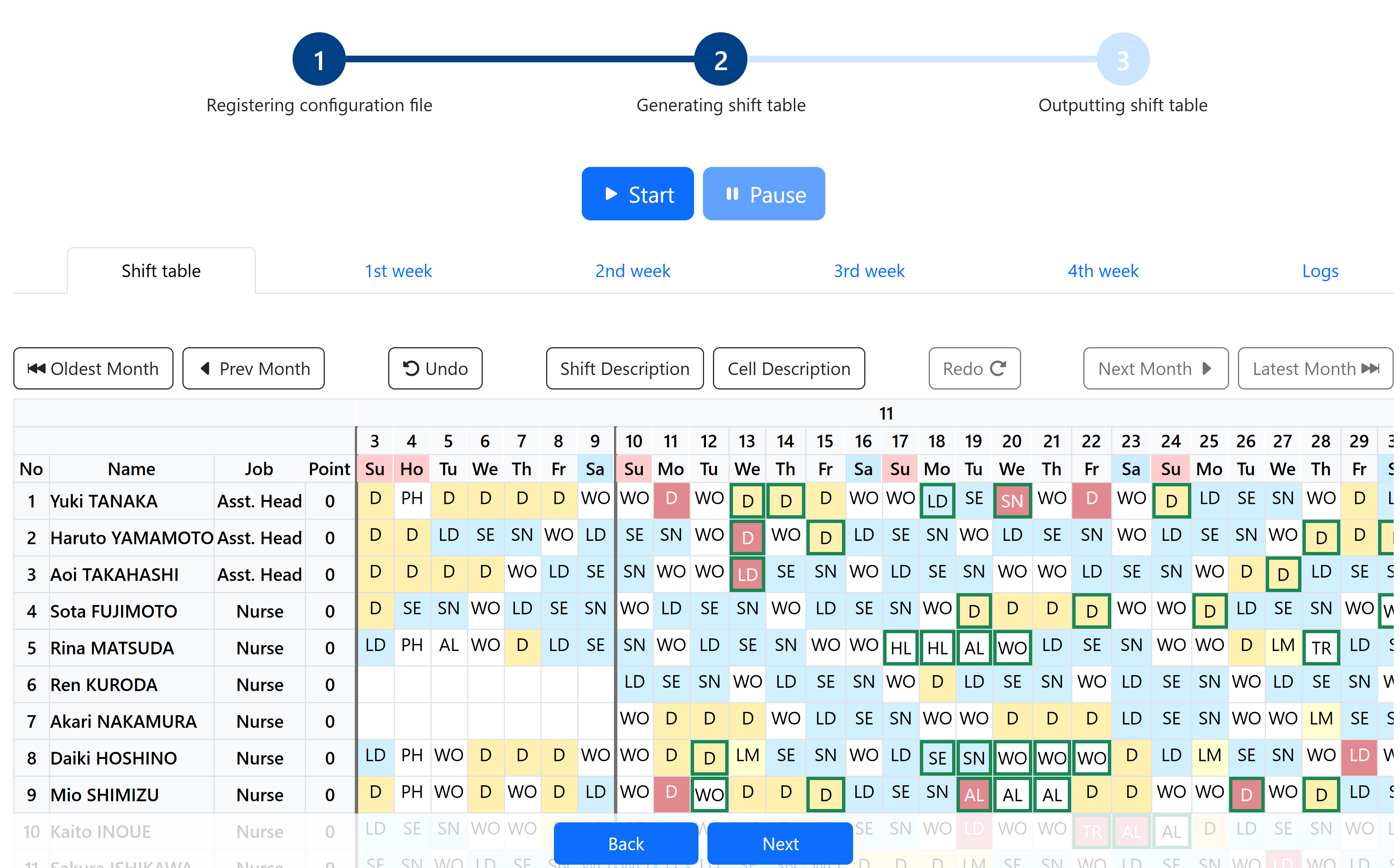}
  \caption{Schedule generation interface in \sysname.}\label{fig:aspital}
\end{figure}
Based on the NSP model presented in the previous section, we developed
\sysname, a system for nurse schedule generation.
\sysname\ is provided as a web service, allowing head nurses to generate and
modify schedules through a browser interface.\footnote{Only the command-line
version is available in our public repository.} Fig.~\ref{fig:aspital} shows an
example of its schedule generation interface.
In the backend, it utilizes the ASP solver \clingo\ to solve the NSP and updates
the schedule immediately whenever a solution changes.
A key feature of \sysname\ is that it not only generates schedules automatically
but also provides tools to assist head nurses in manual adjustments.
Further details are given below.
The system is currently being trialed in six wards of our hospital.

The automatically generated schedule does not always fully satisfy all head
nurse requirements due to the difficulty of extracting their implicit knowledge,
often resulting in an incomplete schedule.
Additionally, reviewing the schedule is a time-consuming and labor-intensive
task, yet its burden is often underestimated, making the execution of the
iterative modeling refinement cycle challenging in real-world operations. Given
that the generated schedule should already meet most of the head nurse’s
requirements, modifying only the unacceptable parts is a more practical
approach.
Furthermore, even after a schedule is finalized, rescheduling is often required
due to unforeseen circumstances, such as sudden absences caused by illness.
Therefore, providing a mechanism to support schedule modifications is essential
in practice.

\begin{figure}[tb]
  \centering
  \begin{subfigure}{0.48\textwidth}
    \includegraphics[width=\textwidth]{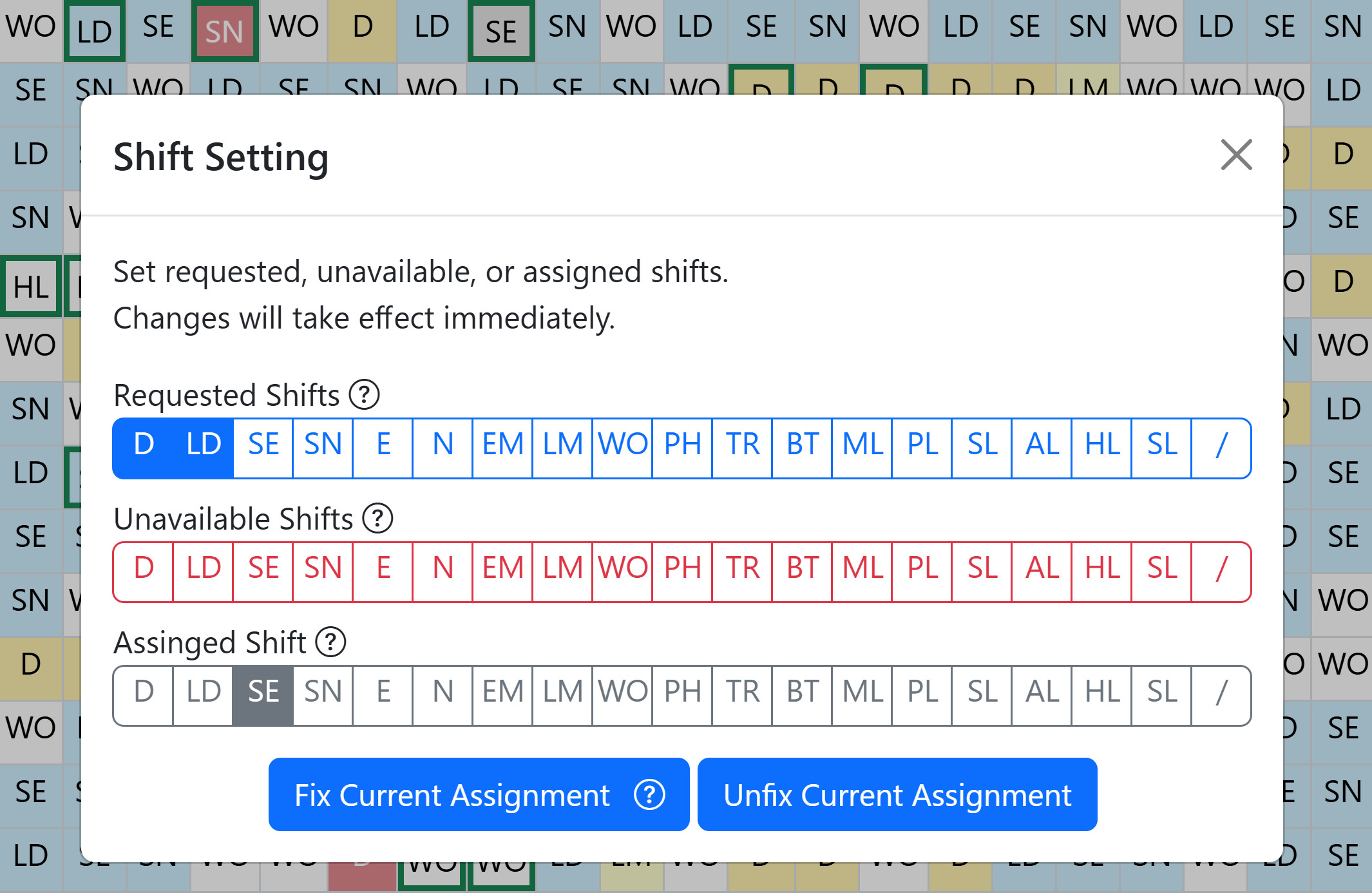}
    \subcaption{Manual shift modifications}
  \end{subfigure}
  \begin{subfigure}{0.48\textwidth}
    \includegraphics[width=\textwidth]{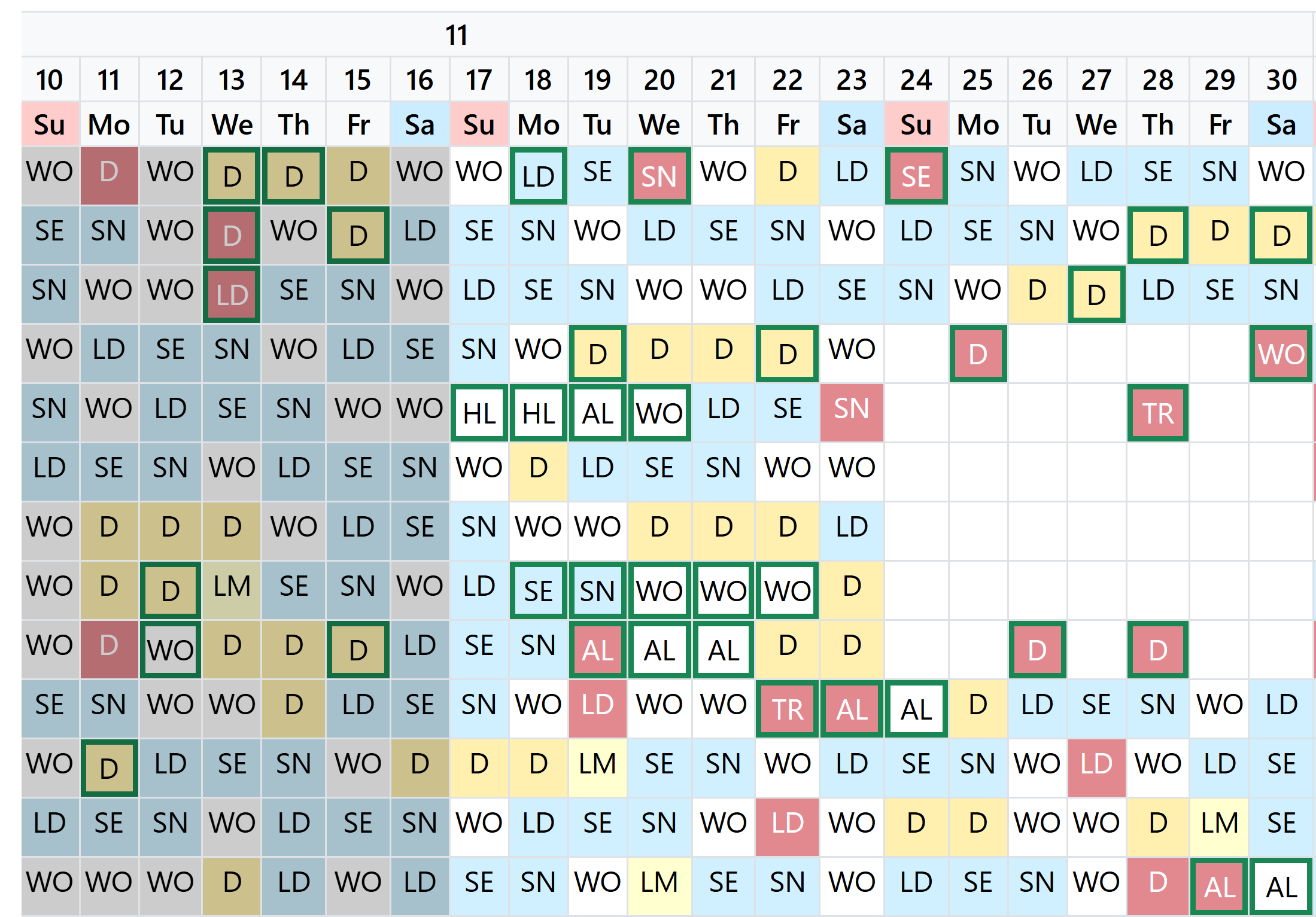}
    \subcaption{Fixing and clearing shift assignments}
  \end{subfigure}
  \caption{Shift editing interface in \sysname}\label{fig:shift-editing}
\end{figure}

To address these challenges, \sysname\ provides various features for schedule
modifications, enabling head nurses to make targeted adjustments efficiently.
If a schedule generated by \sysname\ contains unintended shift assignments, they
can be manually modified to specific shifts. Alternatively, users can specify
lists of desired and undesired shifts, which are implemented by adding the
\verb|pos_request| and \verb|neg_request| predicates as facts. Since \sysname\
is provided as a web service, all modifications can be made directly through a
web browser (see Fig.~\ref{fig:shift-editing} (a)).
Additionally, shift assignments can be fixed, ensuring they remain unchanged in
subsequent optimization iterations. For example, manually modified shifts can be
fixed, or past shifts can be preserved during rescheduling. Conversely, parts of
the schedule can be cleared and reconstructed as needed (see
Fig.~\ref{fig:shift-editing} (b) for an example where the first week's
assignments are fixed while some assignments in the third week are cleared).

A key feature of \sysname\ is its handling of non-modifiable parts of the
schedule. If these parts do not pose any issues, preserving the existing
schedule as much as possible is preferable. As reviewing a schedule can be
burdensome, minimizing unnecessary changes reduces this effort.
Additionally, in rescheduling scenarios, significant modifications to shift
assignments may require confirmation from nurses, making it preferable to
maintain the current schedule whenever possible.
This problem, known as the \emph{Minimal Perturbation Problem} (MPP, hereafter
referred to as MP), aims to make minimal changes to an existing solution while
adapting to new constraints or environmental changes. Traditionally, MP has
been studied within CSPs \citep{sakkout1998,bartak2003}, but its relevance has
recently received more attention \citep{aldoma18b,wickert2019}.
A common approach to MP is incorporating a minimal perturbation term
into the objective function to keep deviations from the initial solution as
small as possible.
In this approach, the trade-off between resolving constraint violations and
minimizing solution changes must be carefully adjusted by tuning the priority of
perturbation minimization or, if incorporated into a single objective function,
by adjusting its weight.

In this study, we apply the \emph{Large Neighborhood Prioritized Search} (LNPS;
\citealp{suinnascsotaba24b}) method to generate modified schedules.
As demonstrated by the experimental results in the next section, LNPS achieves
faster solution improvement than MP in our NSP setting, making it more
suitable for interactive use.
LNPS was originally proposed to efficiently solve combinatorial optimization
problems in ASP. It extends the \emph{Large Neighborhood Search} (LNS) framework
by replacing fixed solution segments with priority-based search, enabling a more
flexible and efficient exploration of feasible solutions. In standard LNS, fixed
segments heavily influence the choice of destruction operators, making it
difficult to guarantee optimality. In contrast, LNPS allows prioritized search
over all variable assignments, facilitating a broader and more effective search.
Prioritized search refers to a method that controls the search process by
preferentially selecting specific value assignments. In LNPS, the value
assignments of the non-disrupted part of the solution are prioritized. However,
this prioritization does not mean the assignments are fixed, as in LNS; if
necessary to satisfy constraints, alternative values may be selected. In other
words, LNPS makes certain values more likely to be chosen but does not enforce
that they are always assigned. This prioritized search is implemented using
the \verb|#heuristic| statement in the ASP solver \clingo, which is used to set
search priorities.

We utilize LNPS for interactive nurse schedule modifications. However, our
approach differs from standard LNPS in the following aspects:
(1)     LNPS periodically restarts the search process. While our method also
        performs periodic restarts, it additionally allows the head nurse to
        manually pause the search, apply modifications, and then restart it. We
        refer to the former as \emph{automatic} restart and the latter as
        \emph{manual} restart. Since \sysname\ updates and displays solutions
        immediately on the web interface, the head nurse can monitor progress
        and pause the search when the solution stabilizes.
(2)     In LNPS, an appropriate destruction operator must be designed for each
        optimization problem. This operator is executed at every automatic
        restart, repeatedly performing destruction and reconstruction to explore
        solutions. In our approach, the head nurse manually selects the parts to
        be destroyed during a manual restart, which corresponds to clearing
        certain shift assignments. No destruction is performed during an
        automatic restart.\footnote{Destruction could be incorporated into
        automatic restarts, but an effective destruction operator has not yet
        been designed for our NSP model.}
(3)     As in LNS, part of the solution can be fixed.

During a manual restart, parts of the solution that are neither fixed nor
cleared by the head nurse become the focus of prioritized search. However, if an
automatic restart occurs in a subsequent search, priorities are reassigned based
on the current variable assignments, and the search resumes accordingly. Thus,
minimal perturbation is not guaranteed, but solution optimality is maintained.

\begin{lstlisting}[float,language=clingof-color,frame=lines,aboveskip=10pt,caption={ASP Encoding for Prioritized Search},label={lst:prioritized}]
#heuristic ext_assigned(N,D,S) : prioritized(ext_assigned(N,D,S)). [1,true]
:- fixed(ext_assigned(N,D,S1)), ext_assigned(N,D,S2), S1 != S2.
\end{lstlisting}
Listing~\ref{lst:prioritized} illustrates the ASP encoding used to implement
prioritized search.
In both automatic and manual restarts, \sysname\ assigns priorities to shift
assignments by adding the fact \texttt{prioritized\allowbreak(\mbox{ext_assigned}(N,D,S))} and
designates fixed shift allocations using the fact
\texttt{fixed(\mbox{ext_assigned}\allowbreak(N,D,S))}.\footnote{Since these predicates must be
asserted and retracted at each restart, \sysname\ utilizes the multi-shot ASP
solving of \clingo\ to manage them dynamically. For details, see
\citep{suinnascsotaba24b}.}
The first line in Listing~\ref{lst:prioritized} specifies that if the fact
\verb|prioritized(ext_assigned(N,D,S))| exists, then \verb|ext_assigned(N,D,S)|
is heuristically prioritized and assigned as true. This is achieved using the
\verb|#heuristic| directive in the ASP solver \clingo.\footnote{To
enable heuristic-driven search in \clingo, the option
\texttt{--heuristic=Domain} must be specified.}
The directive \verb|[1,true]| assigns a priority of \verb|1| to the positive
literal \verb|ext_assigned(N,D,S)|. Since the default priority is \verb|0|, the
solver prefers assigning this literal to true over others with default priority.
The second line enforces an integrity constraint that prevents modifications to
fixed shifts.

\section{Experiments}\label{sec:experiments}

In this section, we compare LNPS and MP using representative NSP instances from
our hospital to demonstrate that LNPS is a suitable search strategy for schedule
adjustments.

The evaluation instances cover a 28-day scheduling period with nurse counts of
10, 20, 30, 40, and 50. For each setting, 10 instances were generated using
different random seeds, totaling 50 instances.
The objective function follows a four-tier priority structure: (1) staff
preferences, (2) inter-shift constraints to prevent invalid assignments, (3)
daily staffing and shift frequency constraints, and (4) all remaining
constraints with the lowest priority.
Requested shifts, both desired and undesired shifts, appear in 10\% of the
schedule cells and are treated as hard constraints. As a result, 60\% of
instances are initially unsatisfiable, requiring relaxation. In such cases, hard
constraints are softened with a priority higher than that of the original soft
constraints, while maintaining the same hierarchical order.
These instances reflect the typical NSP in our hospital. The instance data and
instance generation script are publicly available in our repository.

To evaluate schedule reconstruction performance, an initial solution was
generated for each instance within a one-hour time limit.
Subsequently, 5\% additional requested shifts were introduced, and evaluation
was conducted under three modification scenarios:
\begin{enumerate}
  \item Entire set reconstructed: The initial solution is completely discarded,
        and a new schedule is generated from scratch. This serves as a baseline
        for comparing search strategies without relying on prior assignments.
  \item First half retained, second half reconstructed: The first half of the
        initial solution is preserved, while the second half is cleared and
        reconstructed.
  \item Entire set retained: The entire initial solution is retained.
\end{enumerate}
In scenarios 2 and 3, requested shifts are added even to the retained portions,
potentially causing constraint violations, which may require modifications to
the initial solution to ensure feasibility.

In the evaluation experiments, we compared three search strategies: LNPS, MP,
and MP with Initial-value Search (MP+IS).
LNPS employs a restart mechanism where, if no improved solution is found
        within a given time interval $t$ ($t \in \{10,30,60\}$), the search is
        automatically restarted. Each configuration is denoted as LNPS-$t$
        based on the selected time threshold.
MP minimizes modifications to the initial solution by introducing a
        penalty term into the objective function:
        \begin{lstlisting}[language=clingof-color,numbers=none]
  #minimize { 1@mp_priority,N,D,S :
      prioritized(ext_assigned(N,D,S)), not ext_assigned(N,D,S) }.
        \end{lstlisting}
        where \verb|mp_priority| determines the priority of this objective
        function. In our experiments, we set it to  the highest, middle, and
        lowest priority levels, corresponding to MP-High, MP-Mid, and MP-Low,
        respectively.
MP+IS extends MP by incorporating initial-value-based search, where
        retained assignments from the initial solution guide the search.
These retained values are assigned using \clingo’s \verb|#heuristic|
        directive, as shown in Listing~\ref{lst:init-sign}.
In this encoding, \verb|[10, init]| increases the likelihood of
        selecting \verb|ext_assigned(N,D,S)| in the early search
        stages.\footnote{The value 10 corresponds to the default initial score
        in the VSIDS heuristic. Preliminary tests with values 1, 10, and 100
        showed no significant differences.} The second line, \verb|[1, sign]|,
        instructs the solver to first attempt assigning the variable to true
        once selected as a decision variable.
Similar to MP, we define three variations: MP-IS-High, MP-IS-Mid, and
        MP-IS-Low.

\begin{lstlisting}[float,language=clingof-color,frame=lines,aboveskip=10pt,caption={ASP Encoding for Heuristic Assignment Specification},label={lst:init-sign}]
#heuristic ext_assigned(N,D,S) : prioritized(ext_assigned(N,D,S)). [10,init]
#heuristic ext_assigned(N,D,S) : prioritized(ext_assigned(N,D,S)). [1,sign]
\end{lstlisting}

The performance of each search strategy was evaluated under time limits of 60
and 3600 seconds. The 60-second limit represents a real-time usage scenario,
while the 3600-second limit corresponds to final schedule generation. All
experiments were conducted on a system running Red Hat Enterprise Linux 8 with
an Intel Xeon Platinum 8480+ processor and 512~GiB of RAM.
The solver script was written in Python and executed \clingo~5.7.1 via its Python
API, using the trendy search strategy optimized for industrial instances. The
solver ran in single-threaded mode, and each instance was tested three times.
Additional experimental results, including extended scenarios, alternative
search strategies, and preliminary experiments, are provided in our repository’s
\texttt{comparison} directory.

\begin{figure}[htbp]
  \centering
  \begin{subfigure}{\textwidth}
    \begin{subfigure}{0.49\textwidth}
      \includegraphics[width=\textwidth]{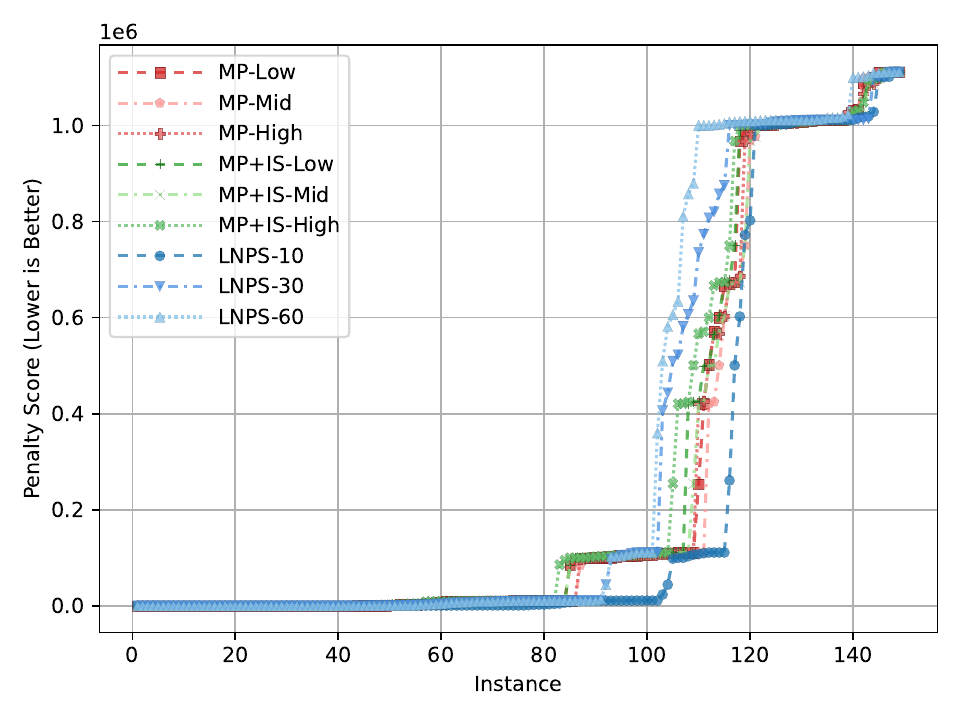}
    \end{subfigure}
    \hfill
    \begin{subfigure}{0.49\textwidth}
      \includegraphics[width=\textwidth]{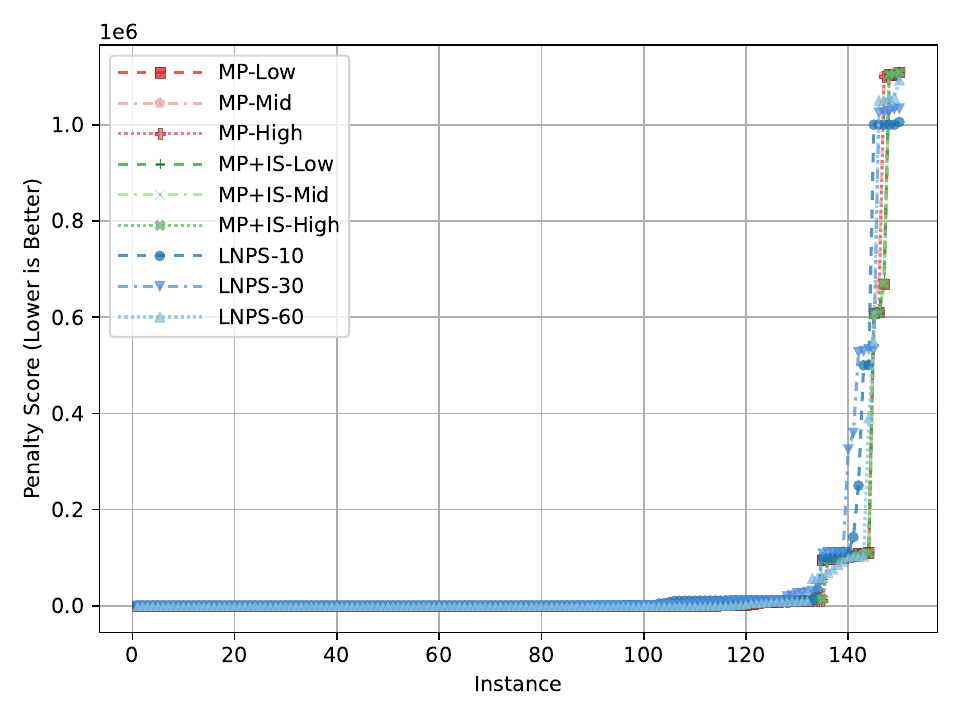}
    \end{subfigure}
  \end{subfigure}
  \caption{Entire set reconstructed (Left: 60-second limit, Right: 3600-second limit)}
  \label{fig:entire-reconstruction}
\end{figure}

Fig.~\ref{fig:entire-reconstruction} shows the results of the entire shift
reconstruction scenario as a cumulative performance plot (cactus plot). Each
search strategy's results are plotted in ascending order of the objective
function value.
Since our NSP is a prioritized multi-objective optimization problem, we apply
scalarization for visualization. The aggregated objective value is computed as
follows:
\begin{displaymath}
  F = \sum_{i=1}^{n} w_i f'_i, \quad f'_i = \frac{f_i - \min f_i}{\max f_i - \min f_i}, \quad w_i = \beta^{(n - i)}
\end{displaymath}
where $n$ is the number of objective functions, and $i$ represents the priority
ranking.
Here, $f'_i$ is the min-max normalized objective value, ensuring comparability
across different scales. The weight parameter $\beta$ is set to 10 to strongly
prioritize higher-ranked objectives while preserving the influence of
lower-priority ones.

\begin{figure}[htbp]
  \centering
  \begin{subfigure}{\textwidth}
    \begin{subfigure}{0.49\textwidth}
      \centering
      \includegraphics[width=\textwidth]{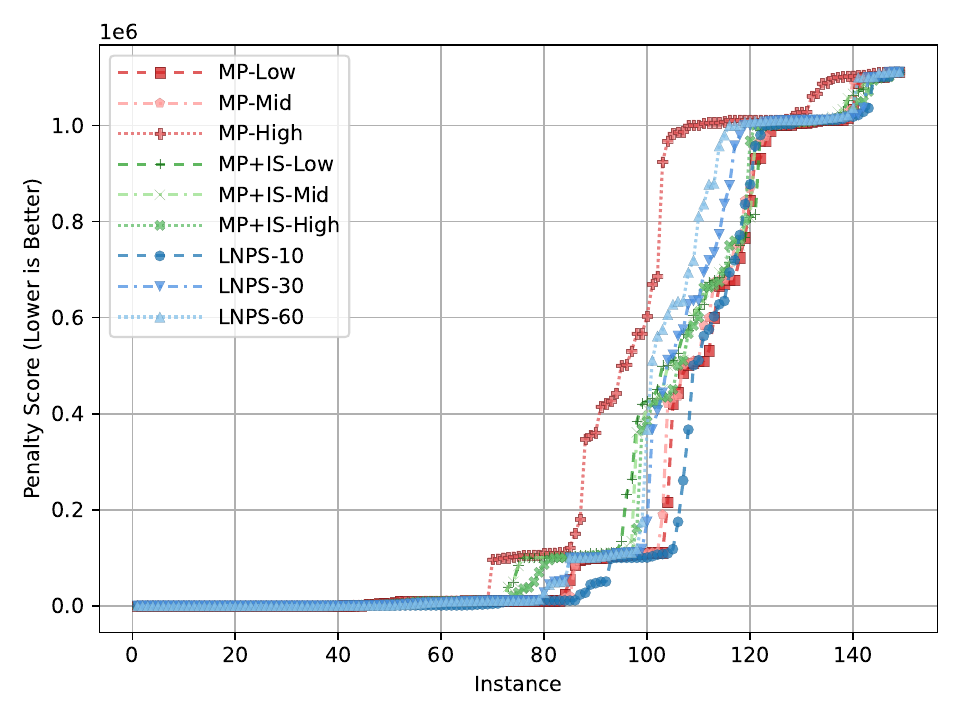}
    \end{subfigure}
    \hfill
    \begin{subfigure}{0.49\textwidth}
      \centering
      \includegraphics[width=\textwidth]{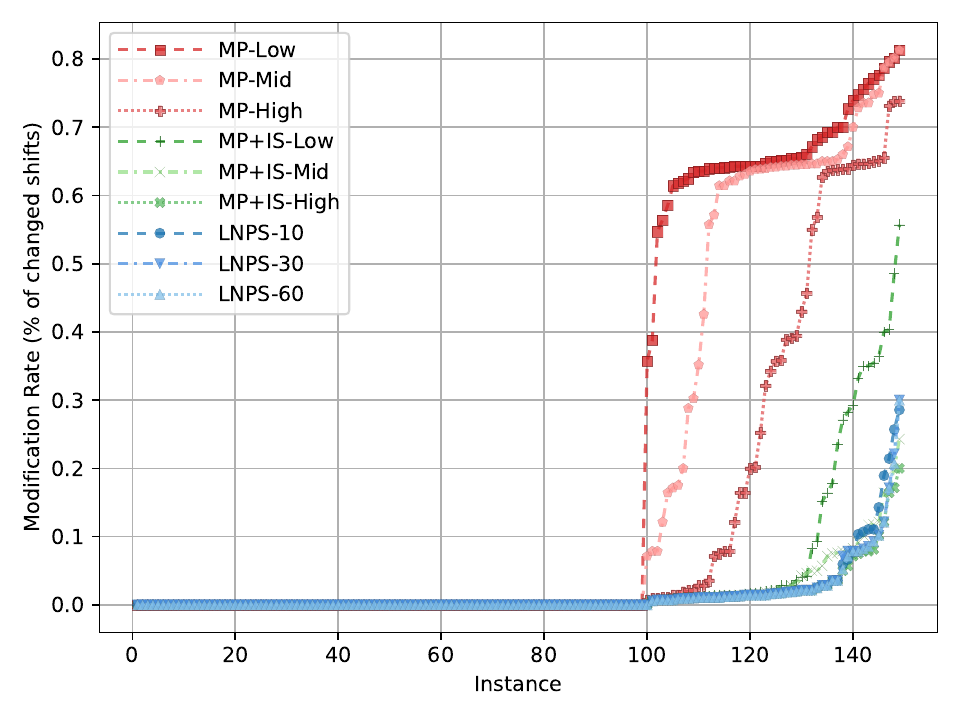}
    \end{subfigure}
    \subcaption{First half retained, second half reconstructed (Left: Penalty score, Right: Modification rate)}
  \end{subfigure}
  \smallskip
  \begin{subfigure}{\textwidth}
    \begin{subfigure}{0.49\textwidth}
      \centering
      \includegraphics[width=\textwidth]{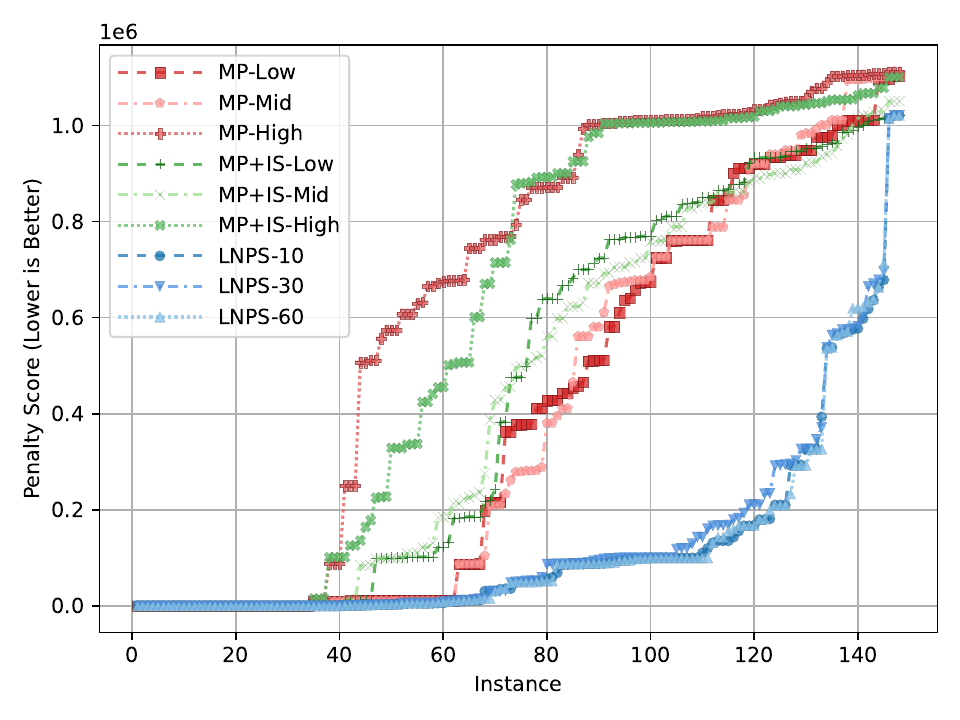}
    \end{subfigure}
    \hfill
    \begin{subfigure}{0.49\textwidth}
      \centering
      \includegraphics[width=\textwidth]{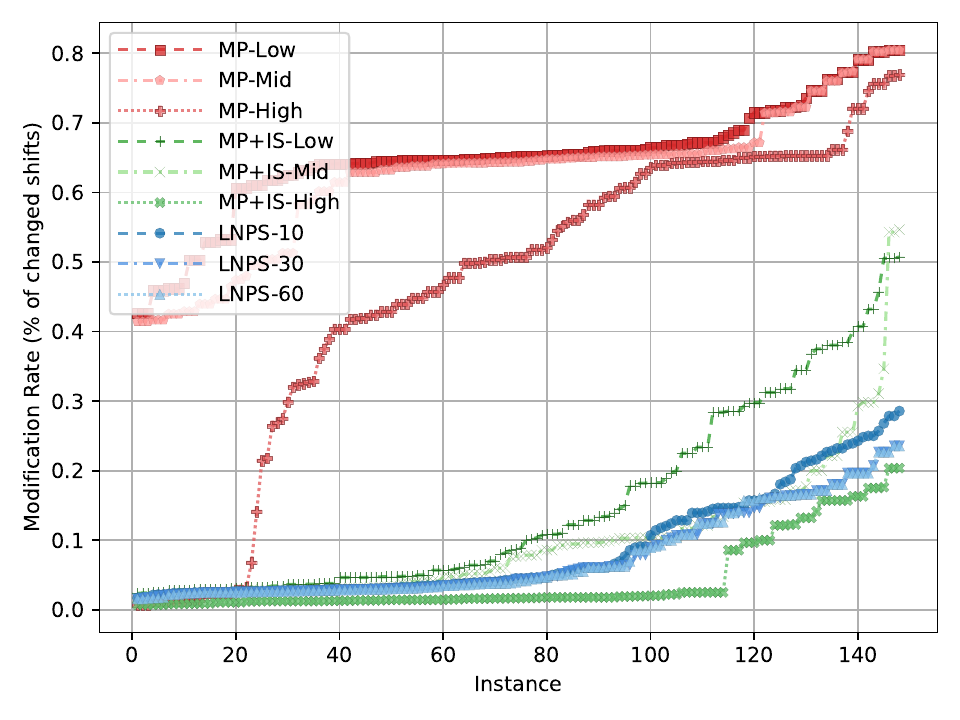}
    \end{subfigure}
    \subcaption{Entire set retained (Left: Penalty score, Right: Modification rate)}
  \end{subfigure}
  \caption{Comparison of search strategies under a 60-second time limit}
  \label{fig:60s-limit}
\end{figure}

\begin{figure}[htbp]
  \centering
  \begin{subfigure}{\textwidth}
    \begin{subfigure}{0.49\textwidth}
      \centering
      \includegraphics[width=\textwidth]{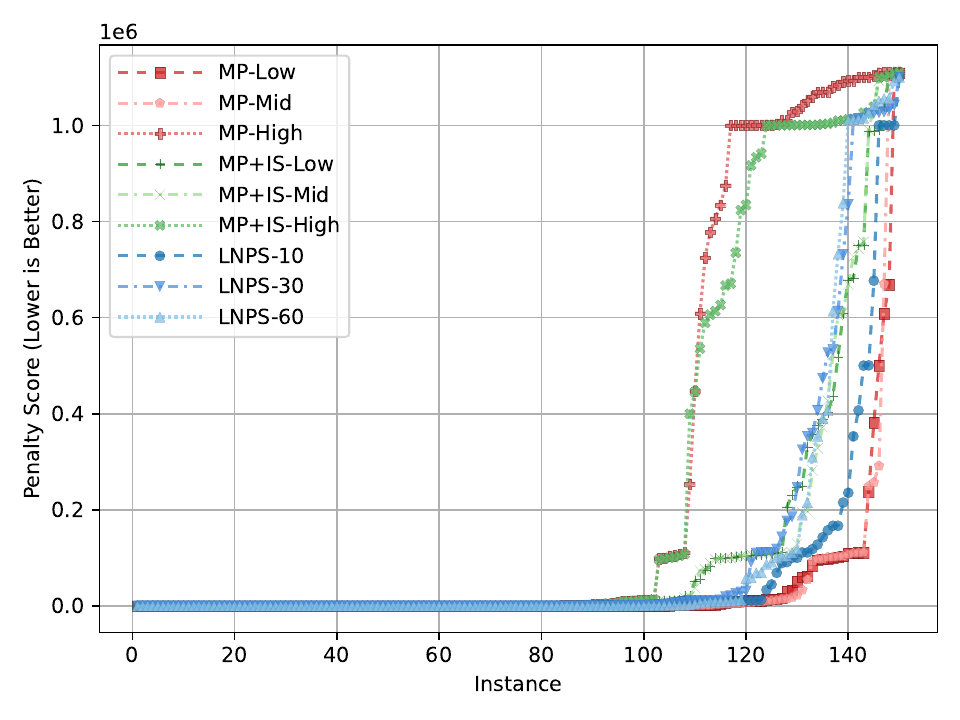}
    \end{subfigure}
    \hfill
    \begin{subfigure}{0.49\textwidth}
      \centering
      \includegraphics[width=\textwidth]{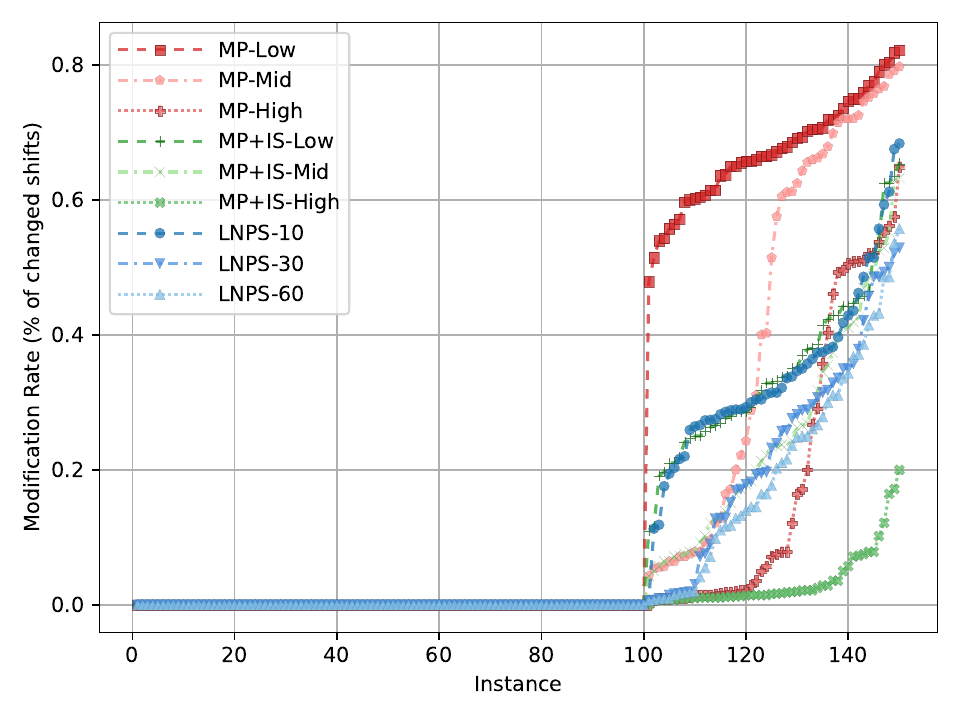}
    \end{subfigure}
    \subcaption{First half retained, second half reconstructed (Left: Penalty score, Right: Modification rate)}
  \end{subfigure}
  \smallskip
  \begin{subfigure}{\textwidth}
    \begin{subfigure}{0.49\textwidth}
      \centering
      \includegraphics[width=\textwidth]{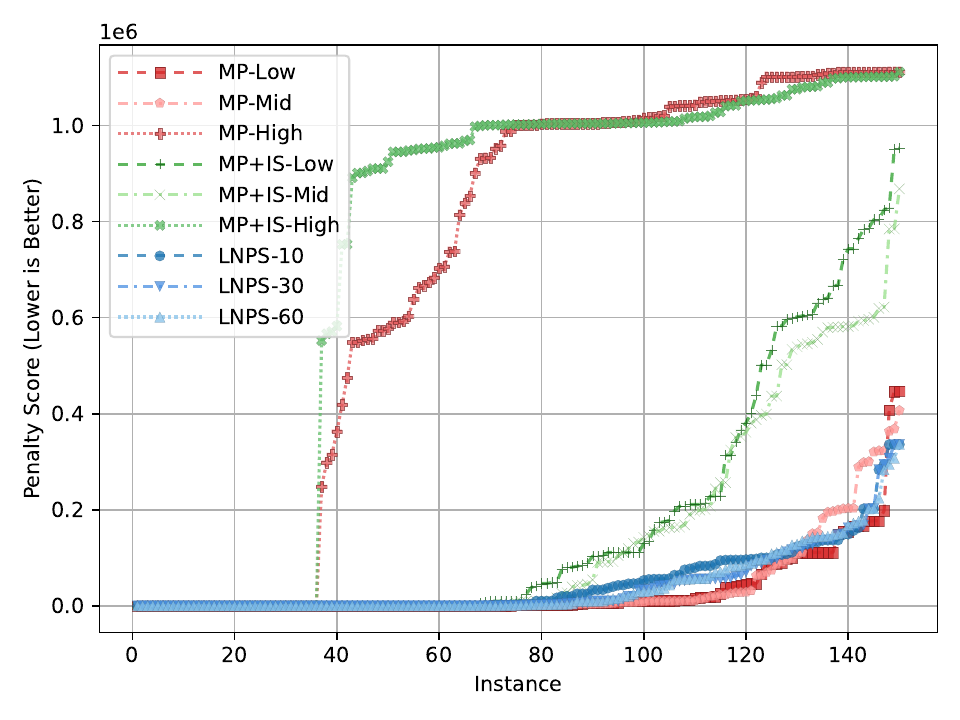}
    \end{subfigure}
    \hfill
    \begin{subfigure}{0.49\textwidth}
      \centering
      \includegraphics[width=\textwidth]{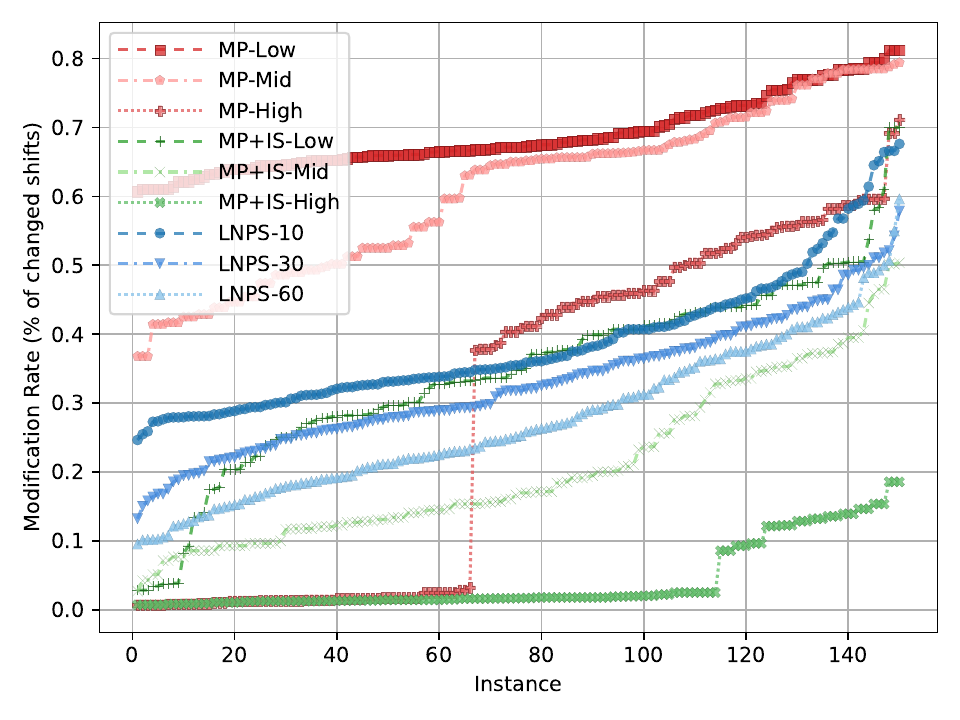}
    \end{subfigure}
    \subcaption{Entire set retained (Left: Penalty score, Right: Modification rate)}
  \end{subfigure}
  \caption{Comparison of search strategies under a 3600-second time limit}
  \label{fig:3600s-limit}
\end{figure}

When reconstructing the entire shift schedule
(Fig.~\ref{fig:entire-reconstruction}), all search strategies perform similarly,
indicating that there is little difference in schedule generation from scratch.
However, when part of the initial solution is retained, performance varies
across strategies. Figs.~\ref{fig:60s-limit} and \ref{fig:3600s-limit} show the
results for 60-second and 3600-second time limits, respectively.

As shown in Fig.~\ref{fig:60s-limit}, LNPS improves the objective function
faster while making fewer modifications to the initial solution. This advantage
stems from LNPS avoiding additional constraints for minimal perturbations,
thereby reducing computational overhead.
However, with longer computation times (Fig.~\ref{fig:3600s-limit}), LNPS
exhibits an increasing number of modifications, eventually reaching levels
similar to MP+IS-Low and MP+IS-Mid. This occurs because LNPS automatically
restarts the search using the latest assignment as a new initial solution,
causing modifications to accumulate over time. Nonetheless, this mechanism helps
escape local optima, potentially leading to better solutions.
Fig.~\ref{fig:3600s-limit} further shows that LNPS achieves objective function
values comparable to MP-Low and MP-Mid while maintaining a lower modification
rate.
MP-Low and MP-Mid achieve better objective function values but require more
modifications. In contrast, MP-IS-High reduces modifications but sacrifices
objective function quality. LNPS balances both, optimizing the objective
function while minimizing modifications, making it well-suited for interactive
use due to its faster computation time.


\section{Discussion}\label{sec:discussion}

Our nurse scheduling system at the University of Yamanashi Hospital builds upon established research in
ASP~\citep{dodmar17a,aldoma18b}, CP~\citep{wehefrpo95a,abdsch99a}, and SAT~\citep{kunach08a,wuchch15a}.
Going beyond these foundations,
our field experience revealed specific challenges in the real-world application,
and we outlined effective strategies for their resolution.
To conclude,
let us present some feedback from nurses using \sysname. Positive feedback includes comments from head nurses indicating a
reduced workload in shift scheduling. Additionally, nurses reported an increased
opportunity to work with a diverse range of colleagues and found it easier to
submit leave requests without concern for the head nurse’s burden of adjusting
schedules.
On the other hand, most negative feedback pertained to shortcomings in the
modeling. For instance, some nurses requested the avoidance of particularly
demanding shift patterns, while others suggested a more balanced allocation of
consecutive days off and night shifts. In response to such feedback,
improvements have been made, such as assigning weekly rest days adjacent to
requested leave ($S_8$) and introducing constraints to equalize workload
distribution ($S_9$).
Regularly incorporating feedback from both head nurses and staff nurses remains
essential for further refining the modeling and enhancing the system’s
effectiveness.

\bibliographystyle{eptcs}

\end{document}